\documentclass[letterpaper]{article} 
\usepackage{aaai25}  
\usepackage{times}  
\usepackage{helvet}  
\usepackage{courier}  
\usepackage[hyphens]{url}  
\usepackage{graphicx} 
\usepackage{natbib}  
\usepackage{caption} 
\usepackage{algorithm}
\usepackage{algorithmic}
\usepackage[inline]{enumitem}

\usepackage{amssymb}
\usepackage{amsthm}

\newtheorem{property}{Property}
\usepackage{multirow}

\usepackage{newfloat}
\usepackage{listings}
\DeclareCaptionStyle{ruled}{labelfont=normalfont,labelsep=colon,strut=off} 
\floatstyle{ruled}
\newfloat{listing}{tb}{lst}{}
\floatname{listing}{Listing}
\usepackage{amsfonts}
\usepackage{amsmath}
\usepackage{bm}
\usepackage{subcaption}
\usepackage{bm}
\usepackage{gmverse}
\usepackage[capitalize,noabbrev]{cleveref}
\usepackage{booktabs}

\DeclareMathOperator*{\argmin}{arg\,min}

\title{From Explainability to Interpretability: \\Interpretable Policies in Reinforcement Learning Via Model Explanation}
\author{
    Peilang Li,
    Umer Siddique,
    Yongcan Cao
}
\affiliations{
    Department of Electrical and Computer Engineering \\
    The University of Texas at San Antonio
    \\
    \{peilang.li, muhammadumer.siddique\}@my.utsa.edu, yongcan.cao@utsa.edu

}

\usepackage{bibentry}

\begin{document}

\maketitle

\begin{abstract}
Deep reinforcement learning (RL) has shown remarkable success in complex domains, however, the inherent black box nature of deep neural network policies raises significant challenges in understanding and trusting the decision-making processes. While existing explainable RL methods provide local insights, they fail to deliver a global understanding of the model, particularly in high-stakes applications. To overcome this limitation, we propose a novel model-agnostic approach that bridges the gap between explainability and interpretability by leveraging Shapley values to transform complex deep RL policies into transparent representations. The proposed approach offers two key contributions: a novel approach employing Shapley values to policy interpretation beyond local explanations and a general framework applicable to off-policy and on-policy algorithms.  We evaluate our approach with three existing deep RL algorithms and validate its performance in two classic control environments. The results demonstrate that our approach not only preserves the original models' performance but also generates more stable interpretable policies.
\end{abstract}



\section{Introduction}
{Reinforcement learning (RL) is an important machine learning technique that learns to make decisions with the best outcomes defined by reward functions~\citep{Sutton2018}. Recent advances in RL have shown remarkable performance when integrating RL with deep learning to solve challenging tasks with human-level or superior performance in, e.g., AlphaGo~\citep{SilverSchrittwieserSimonyanAntonoglouHuangGuezHubertBakerLaiBoltonChenLillicrapHuiSifreDriesscheGraepelHassabis17}, Atari games~\citep{MnihKavukcuogluSilverRusuVenessBellemareGravesRiedmillerFidjelandOstrovskiPetersenBeattieSadikAntonoglouKingKumaranWierstraLeggHassabis15}, and robotics~\citep{gu2017deep}. These successes are largely due to the powerful function approximation capabilities of deep neural networks (DNNs), which excel at feature extraction and generalization. However, the use of DNNs also introduces significant challenges as these models are often considered ``black boxes", making them difficult to interpret~\citep{zahavy2016graying}. They are often complex to train, computationally expensive, data-hungry, and susceptible to biases, unfairness, safety issues, and adversarial attacks~\citep{henderson2018deep,wu2024offline,siddique2020learning}. Thus, an open challenge is to provide quantitative explanations for these models such that they can be understood to gain trustworthiness.} 

Explainable reinforcement learning (XRL) has become an emerging topic that focuses on addressing the aforementioned challenges, aiming at explaining the decision-making processes of RL models to human users in high-stakes, real-world applications. XRL employs the concepts of interpretability and explainability, each with a distinct focus. Interpretability refers to the inherent clarity of a model's structure and functioning, often achieved through simpler models like decision trees~\cite{bastani2018verifiable, pmlr-v108-silva20a} or linear functions that make a policy ``self-explanatory"~\cite{hein2018interpretable}. On the other hand, explainability is related to the use of external, post-hoc methods to provide insights into the behavior of a trained model, aiming to clarify, justify, or rationalize its decisions. Examples include employing Shapley values to determine the importance of state features~\citep{beechey2023explaining} and counterfactual states to gain an understanding of agent behavior~\citep{OLSON2021103455}.

While explainability can provide valuable insights that build user trust, we argue that in high-stakes and real-world applications, explainability alone is insufficient. For instance, Shapley values~\citep{shapley1953value}---a well-known explainable model---provide local explanations by assigning numerical values that indicate the importance of individual features in specific states. Although such explanations can help users build trust by aligning with human intuition and prior knowledge when enough states are covered, they fail to enable users to fully reproduce or predict agent behavior. This is because these local explanations do not provide a comprehensive, global understanding of the model’s functionality, leaving critical aspects of the decision-making process in the dark. In contrast, interpretability offers full transparency and intuitive understanding which is essential for critical applications where trust and comprehensibility are essential. However, the trade-off between simplicity and performance in interpretable models often results in reduced model performance.


Despite its limitations, explainability remains a valuable tool for uncovering insights into model behavior. It can facilitate the development of interpretable policies by abstracting key information from explanations and guiding policy formulation.  In this paper, we propose a model-agnostic approach to generate interpretable policies by leveraging insights from explainability techniques in RL environments. This approach aims at balancing transparency and high performance, ensuring that the resulting models are both understandable and effective.

\paragraph{Contributions.}
In this paper, we present a novel approach that bridges the gap between explainable and interpretable reinforcement learning. Our main contribution is the development of an approach that leverages insights from explainable models to derive interpretable policies. In particular, instead of focusing on the local explanations provided by explainable models, the proposed model-agnostic approach aims to achieve highly transparent and interpretable policies without sacrificing model performance. Additional contributions include the application of the new approach to both off-policy and on-policy RL algorithms and the creation of three adaptations to deep RL methods that learn interpretable policies using insights from model explanation. Finally, we evaluate the effectiveness of our framework in two environments to demonstrate its effectiveness in generating interpretable policies.

\section{Related Work}

One popular approach used in explainable artificial intelligence (XAI) is to use Shapley values that provide a quantitative measure of the contributions of features to the output \citep{JMLR:v11:strumbelj10a, vstrumbelj2014explaining}. In~\citep{10.1145/2939672.2939778}, a method, called LIME, was proposed based on local surrogate models that approximate the predictions made by the original model. In~\citep{wachter2017counterfactual}, the counterfactual is introduced into XAI by producing a perturbation input to change the original prediction to study the intrinsic causality of the model. In~\citep{NIPS2017_7062}, the idea of SHAP was proposed to unify various existing feature attribution methods under a single theoretical framework based on Shapley values, providing consistent and theoretically sound explanations for a wide range of machine learning models.

Most existing explainable methods in RL adopt similar concepts from deep learning via framing the observation as input while the action or reward is the output. In~\citep{beechey2023explaining}, on-manifold Shapley values were proposed to explain the value function and policy that offers more realistic and accurate explanations for RL agents.  In~\citep{OLSON2021103455}, the counterfactual state explanations were developed to examine the impact of altering a state image in an Atari game to understand how these changes influence action selection. As RL possesses some unique challenges, such as sequential decision-making under a reward-driven framework, specialized methods have been considered for its explanation. For example, in~\citep{juozapaitis2019explainable}, reward decomposition was proposed to break down a single reward into multiple meaningful components, providing insights into the factors influencing an agent's action preferences. Moreover, understanding the action selection in certain critical states of the entire sequence can enhance user trust \cite{huang2018establishing}. A summary of important yet not similar sets of states (trajectories) can provide a broader and more comprehensive view of agent behavior \cite{10.5555/3237383.3237869}. 

In contrast to the XRL, research in interpretable RL usually focuses on the transparency of the decision-making processes via, e.g., a simple representation of policies that are understandable to non-experts. The corresponding studies can be divided into direct and indirect approaches~\citep{glanois2024survey}. The direct approach aims to directly search a policy in the environment using the policy deemed interpretable by the designer or user. Examples of the direct methods include the use of decision tree  ~\cite{silva2020optimization} or a simple closed-form formula~\cite{hein2018interpretable} to represent the policy. The direct approach usually requires a prior expert knowledge for initialization to achieve good performance, often for small-scale problems. On the other hand, the indirect approach provides more flexibility by employing a two-step process: (1) train a non-interpretable policy with efficient RL algorithms, and (2) convert this non-interpretable policy into an interpretable one. For instance, ~\citet{bastani2018verifiable} proposed VIPER, a method to learn high-fidelity decision tree policies from original DNN policies. Similarly, ~\citet{verma2018programmatically} proposed PIRL, a method that presents a way to transform the neural network policy into a high-level programming language. Our proposed methods can be categorized into indirect interpretable approaches using Shapley values to transform original policies into simpler but rigorous closed-form function policies. Distinguishing ourselves from existing indirect interpretation approaches, we uniquely incorporate the Shapley value explanation method to generate more accurate and generalizable interpretable policy without relying on predefined interpretable structures.

\section{Background}

\subsection{Reinforcement Learning}
In Reinforcement Learning, an agent interact{s} with its environment, which is {modeled} as a Markov Decision Process (MDP) defined by the tuple $(\mathcal{S}, \mathcal{A}, \mathcal{P}, r, \gamma, d_0)$, where $\mathcal{S}$ is the set of states and $\mathcal{A}$ is the set of {possible actions, $\mathcal{P}: \mathcal{S} \times \mathcal{A} \times \mathcal{S} \rightarrow [0, 1]$ is the transition probability function, $r: \mathcal{S} \times \mathcal{A} \rightarrow \mathbb{R}$ is the reward function, $\gamma \in [0, 1]$ is discount factor, and $d_0: \mathcal{S} \rightarrow [0, 1]$ specifies the initial state distribution.} At time step $t$, the agent observes the current state $s_t \in \mathcal{S}$ and performs an action $a_t \in \mathcal{A}$. {In response,} the environment transitions to a new state $s_{t+1} \sim \mathcal{P}(\cdot|s_t, a_t)$ and {provides} a reward $r_{t+1}$. {The agent's} objective is to learn a policy {(i.e., strategy)} $\pi$ that maximizes the expected return $\mathbb{E}_{\pi}[G_t]$, where $G_t = \sum_{n=t}^{\infty }\gamma^{n}r_{n+1}$. {In RL,} polic{ies} can be deterministic $\pi: \mathcal{S} \rightarrow \mathcal{A}$ or stochastic $\pi: \mathcal{S} \times \mathcal{A} \rightarrow [0, 1]$. {consider an} environment {with} $n$ state features{, where} $\mathcal{S} = \mathcal{S}_1 \times ... \times \mathcal{S}_{n}$, and each state can be represented as an ordered set $s = \{s_i|s_i \in \mathcal{S}_i\}_{i=1}^{n}$. {U}sing $N = \{1, ..., n\}$ to represent the {set of all} state features, {a} partial observation of the state can be {denoted} as the ordered set $s_C = \{s_i|i \in C\}$ where $C \subset N$.

\subsection{Shapley Values in Reinforcement Learning}
The \textit{Shapley value}~\cite{shapley1953value} is a method from cooperative game theory that distributes credit for the total value $v(N)$ earned by a team $N$ among its players. {It is defined as:}
\begin{equation} \label{eq:1}
    \phi_{i}(v)=\sum_{C\subseteq N \setminus \left \{ i \right \}} \frac{\left | C \right |!(n-\left | C \right |-1)!}{(n!)}[v(C \cup \left \{ i \right \})-v(C)],
\end{equation}
{where $v(C)$ represents the value generated by a coalition of players $C$. The Shapley value $\phi_i(v)$ is the average marginal contribution of player $i$ when added to all possible coalitions $C$.}

{In RL,} the state features $\{s_1, ..., s_n\}$ {can be treated as players, and the policy output $\pi(s)$ can be viewed as the total value generated by their contributions. To compute the Shapley values of these players, it is essential to define a characteristic function $v(C)$ that reflects the model's output for a coalition of features $s_C \subseteq {s_1, \dots, s_n}$.}


As the trained policy is undefined for partial input $s_C$, it is important to correctly define the characteristic function for accurate Shapley values calculation. Following the \textit{on-manifold} characteristic value function~\citep{frye2021shapley, beechey2023explaining}, we account for feature correlations rather than assuming independence. 

For a deterministic policy $\pi: S \rightarrow A$, which outputs actions, the characteristic function is defined as:
\begin{equation} \label{eq:2}
    v^{\pi}(C) := \pi_{C}(s) = \sum_{s' \in S} p^{\pi}(s'|s_c)\pi(s'),
\end{equation}
where $ s' = s_C \cup s'_{\bar{C}}$ and $p^{\pi}(s'|s_C)$ is the probability of being in state $s'$ given the limited state features $s_C$ is observed following policy $\pi$. Similarly, for a stochastic policy $\pi: S \times A \rightarrow [0, 1]$, which outputs action probabilities, the characteristic function is defined as:
\begin{equation} \label{eq:3}
    v^{\pi}(C) := \pi_{C}(a|s) = \sum_{s' \in S} p^{\pi}(s'|s_c)\pi(a|s') .
\end{equation}

\section{Method} \label{sec:method}
In this section, we present our proposed methods in two main parts. First, Shapley vectors analysis focuses on extracting and capturing the underneath patterns provided by Shapley values. Secondly, interpretable policy formulation focuses on utilizing these patterns to construct interpretable policies with comparable performance. The complete algorithm is provided in~\Cref{alg:shapley_decision_boundary}.

\subsection{Shapley Vectors Analysis}
Given a well-trained policy $\pi(s)$ (deterministic) or $\pi(a|s)$ (stochastic) in RL, Shapley values provide a way to explain the policy's behavior by quantifying the contributions of state features to the RL policy. Following the Shapley values methods~\citep{beechey2023explaining}, we substitute \eqref{eq:2} or \eqref{eq:3} into the Shapley value formula, namely, \eqref{eq:1}, to compute $\phi_i(v^{\pi})$, \textit{i.e.,} the contribution of feature $i$ to the policy under state $s$.

The computed Shapley values $\phi_i(v^{\pi})$ provide insight into how each state feature $i$ influences action selection.  For example, in an environment with two discrete actions, $a_1 = -1$ and $a_2 = 1$. After computing the Shapley value $\phi_i(v^{\pi})$, a positive $\phi_i(v^{\pi})$ indicates that the feature $i$ encourages the selection of $a_2$, while a negative value suggests a preference for $a_1$.  Notably, Shapley values generalize across features; state features contributing equally to a decision will yield identical values, revealing symmetry in policy reasoning. In this paper, we take this property of Shapley values as their generalization ability.

To exploit this generalization, we represent each state $s$ as a Shapley vector composed of contributions from all features given by
\begin{equation}
    \Phi_{s} = (\phi_1, ..., \phi_n).
\end{equation}
This enables us to cluster the states with similar action selection behavior which further gives insights into action-group boundaries.

\begin{algorithm}[t]
\caption{Shapley Vector Decision Boundary Algorithm}
\label{alg:shapley_decision_boundary}
\textbf{Input}: Shapley vectors $(\Phi_{s_1}, \Phi_{s_2}, ..., \Phi_{s_m})$, Original states $(s_1, s_2, ..., s_m)$\\
\textbf{Parameter}: Action numbers $k$\\
\textbf{Output}: Decision Boundary functions $\{f_{ij}\}$ for each pair of actions $(i,j)$
\begin{algorithmic}[1]
\STATE Initialize empty set of boundary points $B = \{\}$
\STATE $A = \{A_1, ..., A_k\} \leftarrow \text{Action KMeans}(\{\Phi_{s_i}\}_{i=1}^m, k)$
\FOR{$i=1$ to $k$}
    \STATE $\bm{\mu}_i \leftarrow \frac{1}{|A_i|}\sum_{\Phi \in A_i}\Phi$
\ENDFOR
\FOR{$i=1$ to $k-1$}
    \FOR{$j=i+1$ to $k$}
        \STATE $X_{ij} \leftarrow \underset{X}{\argmin} (||X - \bm{\mu}_i||^2 - ||X - \bm{\mu}_j||^2)$
        \STATE $B \leftarrow B \cup \{X_{ij}\}$
        \STATE $s_{ij} \leftarrow \phi^{-1}(X_{ij})$ 
    \ENDFOR
\ENDFOR
\FOR{each pair of clusters $(i,j)$}
    \STATE $f_{ij}(s) \leftarrow \text{Regression}(s_{ij})$
\ENDFOR
\STATE \textbf{return} $\{f_{ij}\}$ 
\end{algorithmic}
\end{algorithm}

\subsubsection{Action K-Means Clustering.}
To cluster states based on their Shapley vectors, we employ action K-means clustering. Given a set of states $(s_1, s_2, ..., s_m)$, where each state is represented by a $n$-dimensional Shapley vector $(\phi_1, \phi_2, ..., \phi_n)$, the algorithm partitions these states into $k$ clusters $A = {A_1, A_2, \dots, A_k}$, where $k$ is the number of discrete actions in the environment. The clustering objective is to minimize inter-cluster variance:
\begin{equation}
    \underset{A}{\argmin} \sum_{i=1}^{k}\sum_{\Phi_s \in A_i}\left\| \Phi_s - \bm{\mu}_{i} \right\|^{2},
\end{equation}
where $\bm{\mu}_{i}$ is the centroid of points in $A_i$, usually represented as $\bm{\mu}_{i} = \frac{1}{|A_i|}\sum_{\Phi_s \in A_i}\Phi_s$.

\subsubsection{Boundary Point Identification.}
Once clusters are formed, the boundaries between action regions can be identified using boundary points.  A \textit{boundary point} $X$ exists at the interface of two clusters $A_i$ and $A_j$, where the policy is equally likely to select either action. This condition arises when the policy is not sure which action to take at the current state, and therefore can serve as a boundary decision. Formally, $X$ is found by minimizing the difference between distances to cluster centroids:
\begin{equation}
    \underset{X}{\argmin} \left( ||X - \bm{\mu}_{i}||^2 - ||X - \bm{\mu}_{j}||^2 \right) ,
\end{equation}
where $\bm{\mu}_{i}$ and $\bm{\mu}_{j}$ are the centroid of points in $A_i$ and $A_j$, respectively.

\begin{property}[Existence and Uniqueness of Decision Boundaries]
For a stationary deterministic policy $\pi$ within an MDP, characterized by a fixed state distribution $d_{\pi} (s)$, there exists a unique boundary surface in the Shapley vector space such that:
\begin{enumerate*}[label=(\roman*)]
    \item the boundary separates the Shapley vectors associated with distinct discrete actions, and
    \item the Euclidean distance from any action's Shapley vector to this boundary remains constant across all states under the stationary policy. 
\end{enumerate*}
\end{property}

\begin{proof}
The efficiency property of Shapley values ensures that the sum of contributions from all features equals the difference between the policy's action value for state $s$ and the expected action value across states, \textit{i.e.},
\begin{equation} \label{eq:7}
    \sum_{i=1}^{n} \phi_{i} = \pi(s) - \mathbb{E}_S(\pi(S)).
\end{equation}
For states $s_p$ and $s_q$ that lead to different action selection $\pi(s_p) = a_p$ and $\pi(s_q) = a_q$, where $a_p \neq a_q$, the difference between their action values defines a gap given by
\begin{equation} \label{eq:8}
     \left|\pi(s_p) - \pi(s_q) \right| =  \left| a_p - a_q\right| = \Delta a .
\end{equation}
Given that the policy $\pi$ is stationary with a fixed state distribution $\mu(s)$, the expected action value converges to a fixed scalar value given by
\begin{equation} \label{eq:9}
    \mathbb{E}_{S \sim \mu}[\pi(S)] = \frac{1}{|\mathcal{S}|} \sum_{s \in \mathcal{S}} \pi(s) = \bar{a} .
\end{equation}
By substituting ~\eqref{eq:8} and \eqref{eq:9} into the efficiency property~\eqref{eq:7}, the Shapley value that sums for all states satisfy a gap
\begin{equation}
    \left| \sum_{i=1}^{n} \phi_{i,s_p} - \sum_{i=1}^{n} \phi_{i,s_q}\right| = \Delta a, \forall s_p, s_q \in \mathcal{S} ,
\end{equation}
where  $\pi(s_p) = a_p \neq \pi(s_q) = a_q$. This implies that the gap $\Delta a$ exists between all states with different action selections. Consequently, we defined the boundary surface $\mathcal{B}$ in the Shapley vector space as
\begin{equation}
    \mathcal{B} = \left\{ \vec{v} \in \mathbb{R}^n  \middle\vert\,  \sum_{i=1}^{n} v_i = \bar{a} + \frac{\Delta a}{2} \right\}.
\end{equation}

The distance from any Shapley vector plane $\Phi$ to this boundary surface $\mathcal{B}$ is given by
\begin{equation}
 \text{dist}(\Phi_{s}, \mathcal{B}) = \frac{\sum_{i=1}^{n} \phi_{i} - \sum_{i=1}^{n} v_i}{\sqrt{n}}.   
\end{equation}
Therefore, for all states, $s_p, s_q \in \mathcal{S}$,  the distances from their Shapley vectors to the boundary remain constant:
\begin{equation}
\text{dist}(\Phi_{s_p}, \mathcal{B}) = \text{dist}(\Phi_{s_q}, \mathcal{B})
\end{equation}
This proves the existence and uniqueness of the decision boundary in the Shapley vector space. The constant distance between the boundary surface and Shapley vector plane lays the foundation for an interpretable policy that maps each action region to its corresponding state region. 
\end{proof}


\subsection{Interpretable Policy Formulation}
With the decision boundary point's identification in the Shapley vector space,  the next step is to map it back to the original state space to construct an interpretable policy.

\subsubsection{Inverse Shapley Values.}
To reconstruct the decision boundary in the state space, we model it as the \textit{Inverse Shapley Value Problem} $\phi_{i}^{-1}: \phi_i(v) \rightarrow \{i\}$, where the goal is to recover the original state $s$ corresponding to a given Shapley vector $\Phi_{s}$. We address this problem by systematically storing the original states with their corresponding Shapley value vectors, enabling efficient inverse function operations. It allows us to map Shapley value vectors back to their original states directly, facilitating precise reconstruction of the decision boundary.

\subsubsection{Decision Boundary Regression.}
After the boundary state points ${s_{ij}}$ are discovered using Shapley values, the decision can be drawn accordingly. While a variety of regression techniques can be used, we use linear regression due to its simplicity and interpretability. The resulting boundary functions $f_{ij}$ define the action regions. 

This policy is then reformulated by assigning actions based on the regions characterized by boundary functions. Specifically, for a given state $s$, the action $a$ is determined by the cluster in which $s$ resides relative to $f_{ij}$.


\section{Experiments}
To evaluate the effectiveness of our proposed method, we performed experiments across two classical control environments from Gymnasium~\cite{towers2024gymnasium}: CartPole and MountainCar.  These environments were specifically chosen as they represent an important control problem where policy interpretability is crucial for real-world deployment. To demonstrate the generality of our framework, we applied it to both off-policy and on-policy deep RL algorithms.  Specifically, we applied it to Deep Q-Network (DQN)~\cite{mnih2015human} as an off-policy method, and Advantage Actor-Critic (A2C)~\cite{pmlr-v48-mniha16} and Proximal Policy Optimization (PPO)~\cite{schulman2017proximal} as on-policy methods. Our experimental results demonstrate that the interpretable policies generated by our method perform competitively to those of deep RL algorithms, and also exhibit better stability and broad applicability.

\begin{figure*}[t]
    \centering
    \includegraphics[width=0.9\linewidth]{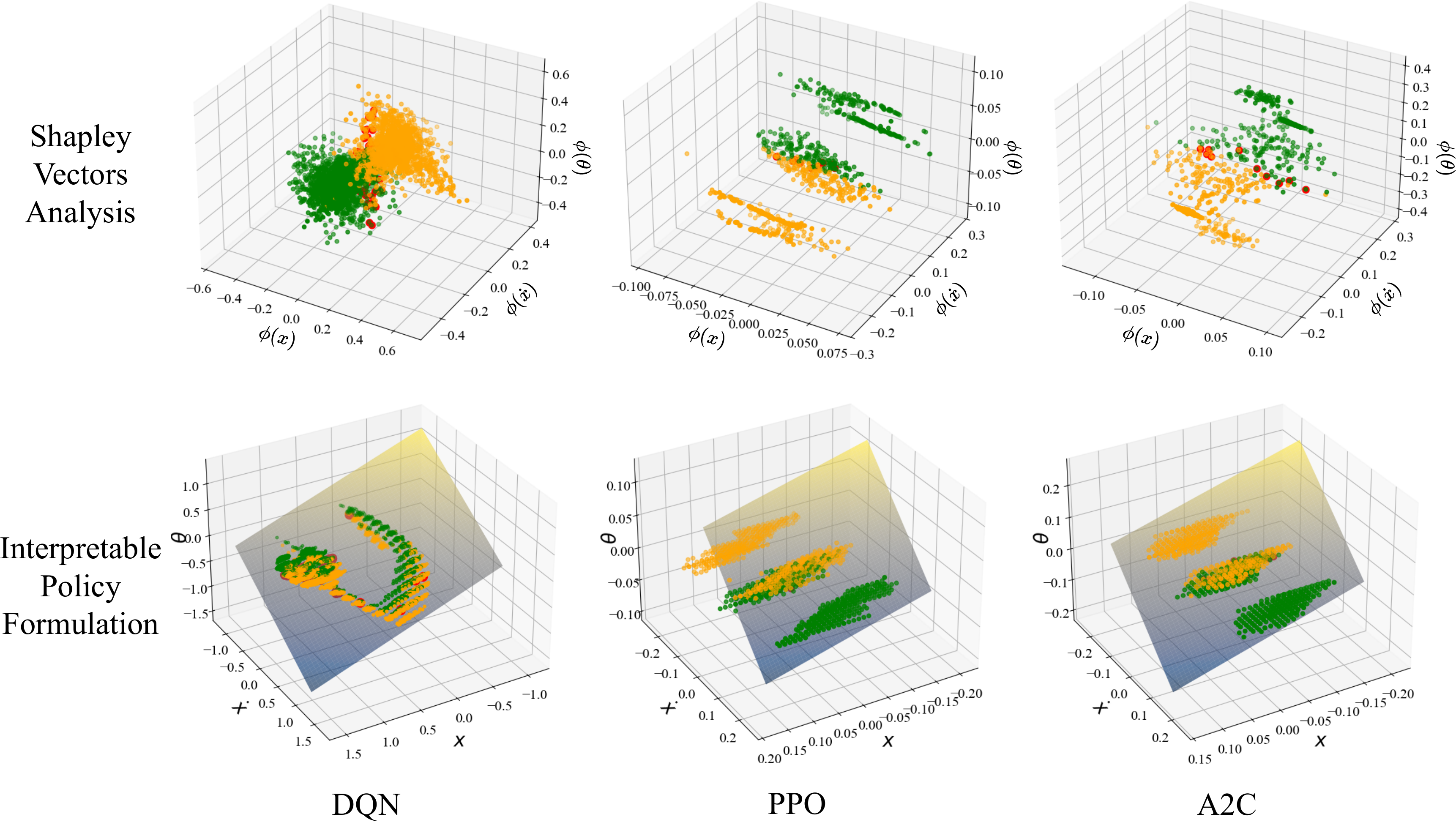}
    \caption{Visualization of Shapley values and interpretable policy formulation in the CartPole. The first row depicts the Shapley value vectors for DQN, PPO, and A2C, with clusters represented in different colors and boundary points highlighted in red. The second row illustrates the corresponding interpretable policy in the original state space, showing decision boundaries that separate the state space into distinct action regions. (Due to the limitations of dimensional plotting, only the first three features $x, \dot{x}, \theta$ are visualized in the figure)}
    \label{fig:cartpole_plot}
\end{figure*}

\renewcommand{\arraystretch}{1.5} 
\begin{table}[t]
\centering
\begin{tabular}{|c|l|}
\hline
\textbf{Algorithm} & \textbf{Decision Boundary} \\ 
\hline \hline
DQN      & $f_{01} = -0.5x-0.687\dot{x}-1.09\theta-\dot{\theta}-0.018$         \\ 
\hline
PPO      & $f_{01} = -0.193x-0.523\dot{x}-\theta-\dot{\theta}+0.0014$         \\ 
\hline
A2C      & $f_{01} = -0.4875x-0.9811\dot{x}-1.09\theta-\dot{\theta}$         \\
\hline
\end{tabular}
\caption{CartPole interpretable policy boundary}
\label{tab:cartpole}
\end{table}

\begin{figure}[t]
    \centering
    \includegraphics[width=1\linewidth]{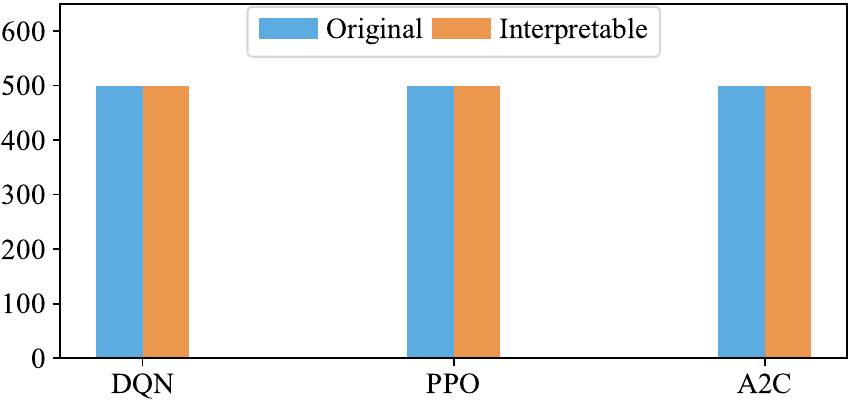}
    \caption{Performances of the interpretable policy with original algorithms---DQN, PPO, A2C in CartPole Environment}
    \label{fig:reward_cartpole}
\end{figure}

\subsection{CartPole}
The CartPole environment is a classic control problem in which an inverted pendulum is placed on the movable cart. The state space in this environment consists of four features: position of cart $x$, velocity of cart $\dot{x}$, angle between the pendulum and the vertical $\theta$, and angular velocity of pendulum $\dot{\theta}$. The action space includes two discrete actions, where the first action $0$ means push the cart to the left, and the second action $1$ means push to the right. A reward of +1 is assigned for each timestep the pole remains upright. The goal in this environment is to balance the pendulum by applying forces in the left and right direction on the cart. 

As explained in method (Section 4), our goal is to obtain an interpretable policy for this problem. To achieve this, we first train three deep RL methods, namely DQN, PPO, and A2C to obtain the optimal policies. Once the models were trained, we evaluated their performance in the CartPole environment and sampled state distributions from 100 trajectories for each algorithm. For each sampled state, we computed the Shapley values of its features using~\Cref{eq:1}. With this step, we construct a Shapley value vector $\Phi_s$ that represents the contribution of state features to this policy's decision. The first row of~\Cref{fig:cartpole_plot}, illustrates the Shapley value vectors for DQN, PPO, and A2C, respectively. Using these Shapley values, we performed k-means clustering on the action space to identify cluster centroids, where each cluster represents a distinct action region. Each cluster is depicted in a different color. We then identified boundary points, which are shown in red in the first row of \Cref{fig:cartpole_plot}. These boundary points indicate the transition between action regions.

\begin{figure*}[t]
    \centering
    \includegraphics[width=0.9\linewidth]{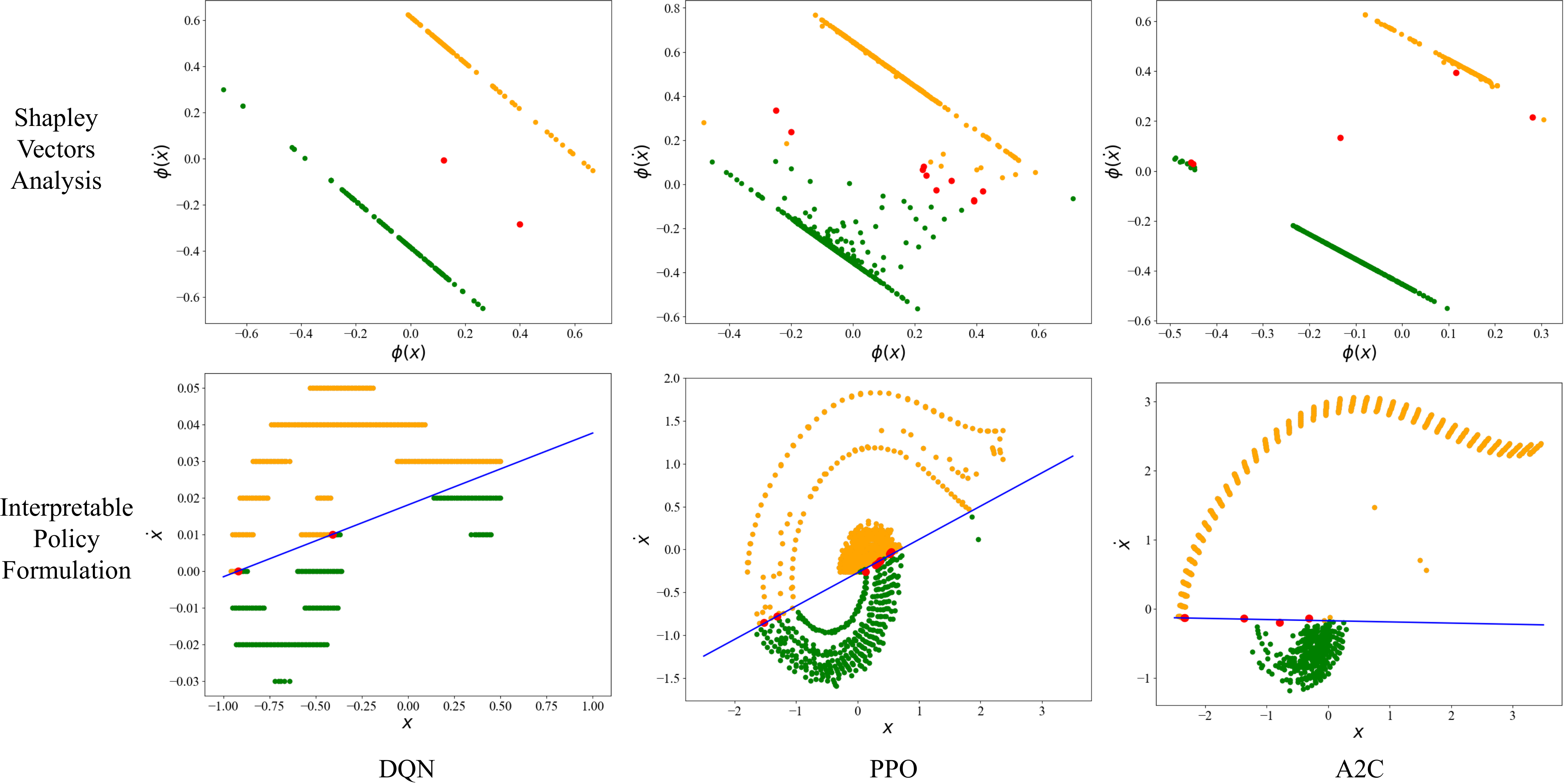}
    \caption{Visualization of Shapley values and interpretable policy formulation in the MountainCar. The first row depicts the Shapley value vectors for DQN, PPO, and A2C, with clusters represented in different colors and boundary points highlighted in red. The second row illustrates the corresponding interpretable policy in the original state space, showing decision boundaries that separate the state space into distinct action regions.}
    \label{fig:mountaincar_plot}
\end{figure*}

Next, we reconstructed the decision boundary in the original state space using the boundary points identified in the Shapley vector space. The second row of \Cref{fig:cartpole_plot} shows these boundaries in the state space for each algorithm. Finally, as described in the methodology, we applied linear regression to derive an interpretable policy $f_{ij}$. The interpretable policies for DQN, PPO, and A2C are summarized in~\Cref{tab:cartpole}. These policies are obtained through their boundaries which separate the states into different action selection regions. In other words, the decision rule for these policies is: if $f_{01}>0$, select action $0$; otherwise, select action $1$. This interpretable policy framework is fully transparent, enabling reproducibility and mitigating risks in high-stakes real-world applications.

To evaluate the performance of the interpretable policies, we tested them alongside the original deep RL policies over 10 episodes. The results, shown in \Cref{fig:reward_cartpole}, demonstrate that the interpretable policies consistently achieved the maximum reward of 500 across all algorithms. This indicates that our method preserves the performance of the original deep RL algorithms while providing interpretability. These results also highlight the generality and model-agnostic nature of the proposed framework.

\subsection{MountainCar}

\begin{table}[t]
\centering
\begin{tabular}{|c|l|}
\hline
\textbf{Algorithm} & \textbf{Decision Boundary}  \\ 
\hline
\hline
DQN      & $f_{01} = 0.013x - \dot{x} +0.0033$      \\ 
\hline
PPO      & $f_{01} = 0.35x - \dot{x} -0.3$           \\ 
\hline
A2C      & $f_{01} = 0.003x - \dot{x} -0.12$       \\
\hline
\end{tabular}
\caption{MountainCar interpretable policy boundary}
\label{tab:mountaincar}
\end{table}

\begin{figure}[t]
    \centering
    \includegraphics[width=1.0\linewidth]{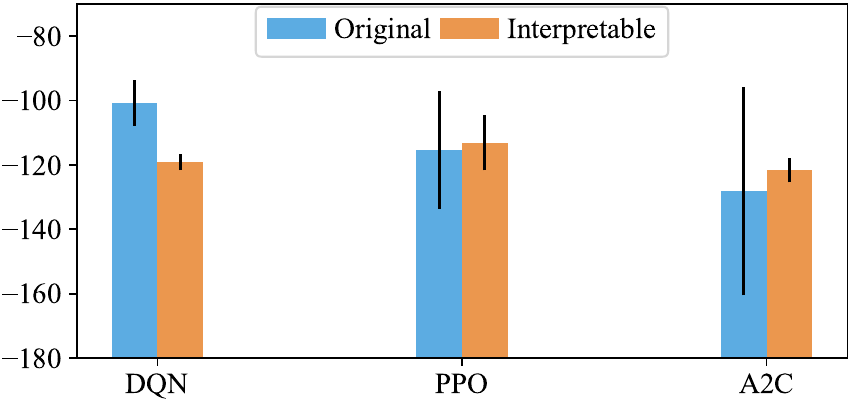}
    \caption{Performances of the interpretable policy with original algorithms—DQN, PPO, A2C in MountainCar Environment.}
    \label{fig:reward_mountaincar}
\end{figure}

The MountainCar environment is another classic control problem where a car is placed at the bottom of a sinusoidal valley. The state space for this environment consists of two features: car position along the x-axis $x$ and the velocity of the car $\dot{x}$. The actions space contains two discrete actions: action $0$ applies left acceleration on the car and action $1$ applies right acceleration on the car. The goal of this environment is to accelerate the car to reach the goal state on top of the right hill. A reward of $-1$ is assigned for each timestep as punishment if the car fails to reach the goal state.

Following the proposed method (section 4), we perform the Shapley vectors analysis in three trained deep RL methods DQN, PPO, and A2C in the MountainCar environment. The result is shown in the first row of~\Cref{fig:reward_mountaincar}. Each cluster represents a distinct action region, distinguished by a unique color and boundary points are highlighted in red. By mapping these boundary points back to the original state space, we constructed the decision boundaries using linear regression, illustrated in the second row of \Cref{fig:mountaincar_plot} as blue lines. The detailed interpretable policies for DQN, PPO, and A2C are in~\Cref{tab:mountaincar} and the decision rule is straightforward: when $f_{01} > 0$, action 0 is chosen, otherwise, action 1 is chosen.

Performance of the interpretable policies alongside the original algorithms was evaluated over 10 episodes, with results presented in Figure~\ref{fig:reward_mountaincar}. Interestingly, interpretable policies derived from PPO and A2C surprisingly outperformed their original algorithms, whereas the interpretable policy generated from DQN experienced a slight performance reduction. A notable observation is that all interpretable policies achieved significantly smaller standard deviations compared to their original counterparts, indicating more stable policy performance. This characteristic is particularly valuable in real-world applications where consistent and predictable behavior is crucial.

\section{Conclusions and Future work}
In this paper, we formalized and addressed the unsolved problem of extracting interpretable policies from explainable methods in RL. 
We propose a novel approach that leverages Shapley values to generate transparent and interpretable policies for both off-policy and on-policy deep RL algorithms.
Through comprehensive experiments conducted in two classic control environments using three deep RL algorithms, we demonstrated that our proposed method achieves comparable performance while generating interpretable and stable policies.

Our future work will include: (1) extending the current approach to continuous action spaces by discretizing the action space, (2) conducting a scalability study of the proposed approach in more complex environments with higher-dimensional state feature spaces, and (3) exploring performance differences across various regression methods.

\section{Acknowledgements}
This work was supported by the Office of Naval Research under Grants N000142412405 and N000142212474.

\bibliography{dai}

\begin{thebibliography}{30}
\providecommand{\natexlab}[1]{#1}

\bibitem[{Amir and Amir(2018)}]{10.5555/3237383.3237869}
Amir, D.; and Amir, O. 2018.
\newblock HIGHLIGHTS: Summarizing Agent Behavior to People.
\newblock In \emph{Proceedings of the 17th International Conference on Autonomous Agents and MultiAgent Systems}, AAMAS '18, 1168–1176. Richland, SC: International Foundation for Autonomous Agents and Multiagent Systems.

\bibitem[{Bastani, Pu, and Solar-Lezama(2018)}]{bastani2018verifiable}
Bastani, O.; Pu, Y.; and Solar-Lezama, A. 2018.
\newblock Verifiable reinforcement learning via policy extraction.
\newblock \emph{Advances in neural information processing systems}, 31.

\bibitem[{Beechey, Smith, and {\c{S}}im{\c{s}}ek(2023)}]{beechey2023explaining}
Beechey, D.; Smith, T.~M.; and {\c{S}}im{\c{s}}ek, {\"O}. 2023.
\newblock Explaining reinforcement learning with shapley values.
\newblock In \emph{International Conference on Machine Learning}, 2003--2014. PMLR.

\bibitem[{Frye et~al.(2021)Frye, de~Mijolla, Begley, Cowton, Stanley, and Feige}]{frye2021shapley}
Frye, C.; de~Mijolla, D.; Begley, T.; Cowton, L.; Stanley, M.; and Feige, I. 2021.
\newblock Shapley explainability on the data manifold.
\newblock In \emph{International Conference on Learning Representations}.

\bibitem[{Glanois et~al.(2024)Glanois, Weng, Zimmer, Li, Yang, Hao, and Liu}]{glanois2024survey}
Glanois, C.; Weng, P.; Zimmer, M.; Li, D.; Yang, T.; Hao, J.; and Liu, W. 2024.
\newblock A survey on interpretable reinforcement learning.
\newblock \emph{Machine Learning}, 1--44.

\bibitem[{Gu et~al.(2017)Gu, Holly, Lillicrap, and Levine}]{gu2017deep}
Gu, S.; Holly, E.; Lillicrap, T.; and Levine, S. 2017.
\newblock Deep reinforcement learning for robotic manipulation with asynchronous off-policy updates.
\newblock In \emph{2017 IEEE international conference on robotics and automation}, 3389--3396. IEEE.

\bibitem[{Hein, Udluft, and Runkler(2018)}]{hein2018interpretable}
Hein, D.; Udluft, S.; and Runkler, T.~A. 2018.
\newblock Interpretable policies for reinforcement learning by genetic programming.
\newblock \emph{Engineering Applications of Artificial Intelligence}, 76: 158--169.

\bibitem[{Henderson et~al.(2018)Henderson, Islam, Bachman, Pineau, Precup, and Meger}]{henderson2018deep}
Henderson, P.; Islam, R.; Bachman, P.; Pineau, J.; Precup, D.; and Meger, D. 2018.
\newblock Deep reinforcement learning that matters.
\newblock In \emph{Proceedings of the AAAI conference on artificial intelligence}, volume~32.

\bibitem[{Huang et~al.(2018)Huang, Bhatia, Abbeel, and Dragan}]{huang2018establishing}
Huang, S.~H.; Bhatia, K.; Abbeel, P.; and Dragan, A.~D. 2018.
\newblock Establishing appropriate trust via critical states.
\newblock In \emph{2018 IEEE/RSJ international conference on intelligent robots and systems (IROS)}, 3929--3936. IEEE.

\bibitem[{Juozapaitis et~al.(2019)Juozapaitis, Koul, Fern, Erwig, and Doshi-Velez}]{juozapaitis2019explainable}
Juozapaitis, Z.; Koul, A.; Fern, A.; Erwig, M.; and Doshi-Velez, F. 2019.
\newblock Explainable reinforcement learning via reward decomposition.
\newblock In \emph{IJCAI/ECAI Workshop on explainable artificial intelligence}.

\bibitem[{Lundberg and Lee(2017)}]{NIPS2017_7062}
Lundberg, S.~M.; and Lee, S.-I. 2017.
\newblock A Unified Approach to Interpreting Model Predictions.
\newblock In Guyon, I.; Luxburg, U.~V.; Bengio, S.; Wallach, H.; Fergus, R.; Vishwanathan, S.; and Garnett, R., eds., \emph{Advances in Neural Information Processing Systems 30}, 4765--4774. Curran Associates, Inc.

\bibitem[{Mnih et~al.(2016)Mnih, Badia, Mirza, Graves, Lillicrap, Harley, Silver, and Kavukcuoglu}]{pmlr-v48-mniha16}
Mnih, V.; Badia, A.~P.; Mirza, M.; Graves, A.; Lillicrap, T.; Harley, T.; Silver, D.; and Kavukcuoglu, K. 2016.
\newblock Asynchronous Methods for Deep Reinforcement Learning.
\newblock In Balcan, M.~F.; and Weinberger, K.~Q., eds., \emph{Proceedings of The 33rd International Conference on Machine Learning}, volume~48 of \emph{Proceedings of Machine Learning Research}, 1928--1937. New York, New York, USA: PMLR.

\bibitem[{Mnih et~al.(2015{\natexlab{a}})Mnih, Kavukcuoglu, Silver, Rusu, Veness, Bellemare, Graves, Riedmiller, Fidjeland, Ostrovski, Petersen, Beattie, Sadik, Antonoglou, King, Kumaran, Wierstra, Legg, and Hassabis}]{MnihKavukcuogluSilverRusuVenessBellemareGravesRiedmillerFidjelandOstrovskiPetersenBeattieSadikAntonoglouKingKumaranWierstraLeggHassabis15}
Mnih, V.; Kavukcuoglu, K.; Silver, D.; Rusu, A.~A.; Veness, J.; Bellemare, M.~G.; Graves, A.; Riedmiller, M.; Fidjeland, A.~K.; Ostrovski, G.; Petersen, S.; Beattie, C.; Sadik, A.; Antonoglou, I.; King, H.; Kumaran, D.; Wierstra, D.; Legg, S.; and Hassabis, D. 2015{\natexlab{a}}.
\newblock Human-level control through deep reinforcement learning.
\newblock \emph{Nature}, 518: 529--533.

\bibitem[{Mnih et~al.(2015{\natexlab{b}})Mnih, Kavukcuoglu, Silver, Rusu, Veness, Bellemare, Graves, Riedmiller, Fidjeland, Ostrovski et~al.}]{mnih2015human}
Mnih, V.; Kavukcuoglu, K.; Silver, D.; Rusu, A.~A.; Veness, J.; Bellemare, M.~G.; Graves, A.; Riedmiller, M.; Fidjeland, A.~K.; Ostrovski, G.; et~al. 2015{\natexlab{b}}.
\newblock Human-level control through deep reinforcement learning.
\newblock \emph{nature}, 518(7540): 529--533.

\bibitem[{Olson et~al.(2021)Olson, Khanna, Neal, Li, and Wong}]{OLSON2021103455}
Olson, M.~L.; Khanna, R.; Neal, L.; Li, F.; and Wong, W.-K. 2021.
\newblock Counterfactual state explanations for reinforcement learning agents via generative deep learning.
\newblock \emph{Artificial Intelligence}, 295: 103455.

\bibitem[{Ribeiro, Singh, and Guestrin(2016)}]{10.1145/2939672.2939778}
Ribeiro, M.~T.; Singh, S.; and Guestrin, C. 2016.
\newblock "Why Should I Trust You?": Explaining the Predictions of Any Classifier.
\newblock In \emph{Proceedings of the 22nd ACM SIGKDD International Conference on Knowledge Discovery and Data Mining}, KDD '16, 1135–1144. New York, NY, USA: Association for Computing Machinery.
\newblock ISBN 9781450342322.

\bibitem[{Schulman et~al.(2017)Schulman, Wolski, Dhariwal, Radford, and Klimov}]{schulman2017proximal}
Schulman, J.; Wolski, F.; Dhariwal, P.; Radford, A.; and Klimov, O. 2017.
\newblock Proximal policy optimization algorithms.
\newblock \emph{arXiv preprint arXiv:1707.06347}.

\bibitem[{Shapley(1953)}]{shapley1953value}
Shapley, L.~S. 1953.
\newblock A value for n-person games.
\newblock \emph{Contribution to the Theory of Games}, 2.

\bibitem[{Siddique, Weng, and Zimmer(2020)}]{siddique2020learning}
Siddique, U.; Weng, P.; and Zimmer, M. 2020.
\newblock Learning fair policies in multi-objective (deep) reinforcement learning with average and discounted rewards.
\newblock In \emph{International Conference on Machine Learning}, 8905--8915. PMLR.

\bibitem[{Silva et~al.(2020{\natexlab{a}})Silva, Gombolay, Killian, Jimenez, and Son}]{pmlr-v108-silva20a}
Silva, A.; Gombolay, M.; Killian, T.; Jimenez, I.; and Son, S.-H. 2020{\natexlab{a}}.
\newblock Optimization Methods for Interpretable Differentiable Decision Trees Applied to Reinforcement Learning.
\newblock In Chiappa, S.; and Calandra, R., eds., \emph{Proceedings of the Twenty Third International Conference on Artificial Intelligence and Statistics}, volume 108 of \emph{Proceedings of Machine Learning Research}, 1855--1865. PMLR.

\bibitem[{Silva et~al.(2020{\natexlab{b}})Silva, Gombolay, Killian, Jimenez, and Son}]{silva2020optimization}
Silva, A.; Gombolay, M.; Killian, T.; Jimenez, I.; and Son, S.-H. 2020{\natexlab{b}}.
\newblock Optimization methods for interpretable differentiable decision trees applied to reinforcement learning.
\newblock In \emph{International conference on artificial intelligence and statistics}, 1855--1865. PMLR.

\bibitem[{Silver et~al.(2017)Silver, Schrittwieser, Simonyan, Antonoglou, Huang, Guez, Hubert, Baker, Lai, Bolton, Chen, Lillicrap, Hui, Sifre, van~den Driessche, Graepel, and Hassabis}]{SilverSchrittwieserSimonyanAntonoglouHuangGuezHubertBakerLaiBoltonChenLillicrapHuiSifreDriesscheGraepelHassabis17}
Silver, D.; Schrittwieser, J.; Simonyan, K.; Antonoglou, I.; Huang, A.; Guez, A.; Hubert, T.; Baker, L.; Lai, M.; Bolton, A.; Chen, Y.; Lillicrap, T.; Hui, F.; Sifre, L.; van~den Driessche, G.; Graepel, T.; and Hassabis, D. 2017.
\newblock Mastering the game of Go without human knowledge.
\newblock \emph{Nature}, 550: 354--359.

\bibitem[{{\v{S}}trumbelj and Kononenko(2010)}]{JMLR:v11:strumbelj10a}
{\v{S}}trumbelj, E.; and Kononenko, I. 2010.
\newblock An Efficient Explanation of Individual Classifications using Game Theory.
\newblock \emph{Journal of Machine Learning Research}, 11(1): 1--18.

\bibitem[{{\v{S}}trumbelj and Kononenko(2014)}]{vstrumbelj2014explaining}
{\v{S}}trumbelj, E.; and Kononenko, I. 2014.
\newblock Explaining prediction models and individual predictions with feature contributions.
\newblock \emph{Knowledge and information systems}, 41: 647--665.

\bibitem[{Sutton and Barto(2018)}]{Sutton2018}
Sutton, R.~S.; and Barto, A.~G. 2018.
\newblock \emph{Reinforcement Learning: An Introduction}.
\newblock The MIT Press, second edition.

\bibitem[{Towers et~al.(2024)Towers, Kwiatkowski, Terry, Balis, De~Cola, Deleu, Goulao, Kallinteris, Krimmel, KG et~al.}]{towers2024gymnasium}
Towers, M.; Kwiatkowski, A.; Terry, J.; Balis, J.~U.; De~Cola, G.; Deleu, T.; Goulao, M.; Kallinteris, A.; Krimmel, M.; KG, A.; et~al. 2024.
\newblock Gymnasium: A standard interface for reinforcement learning environments.
\newblock \emph{arXiv preprint arXiv:2407.17032}.

\bibitem[{Verma et~al.(2018)Verma, Murali, Singh, Kohli, and Chaudhuri}]{verma2018programmatically}
Verma, A.; Murali, V.; Singh, R.; Kohli, P.; and Chaudhuri, S. 2018.
\newblock Programmatically interpretable reinforcement learning.
\newblock In \emph{International Conference on Machine Learning}, 5045--5054. PMLR.

\bibitem[{Wachter, Mittelstadt, and Russell(2017)}]{wachter2017counterfactual}
Wachter, S.; Mittelstadt, B.; and Russell, C. 2017.
\newblock Counterfactual explanations without opening the black box: Automated decisions and the GDPR.
\newblock \emph{Harv. JL \& Tech.}, 31: 841.

\bibitem[{Wu et~al.(2024)Wu, Siddique, Sinha, and Cao}]{wu2024offline}
Wu, M.; Siddique, U.; Sinha, A.; and Cao, Y. 2024.
\newblock Offline Reinforcement Learning with Failure Under Sparse Reward Environments.
\newblock In \emph{2024 IEEE 3rd International Conference on Computing and Machine Intelligence (ICMI)}, 1--5. IEEE.

\bibitem[{Zahavy, Ben-Zrihem, and Mannor(2016)}]{zahavy2016graying}
Zahavy, T.; Ben-Zrihem, N.; and Mannor, S. 2016.
\newblock Graying the black box: Understanding dqns.
\newblock In \emph{International conference on machine learning}, 1899--1908. PMLR.

\end{thebibliography}

\end{document}